# Food safety risk prediction with Deep Learning models using categorical embeddings on European Union RASFF Portal data.

Alberto Nogales[1][¶]*, Rodrigo Díaz Morón[1] and Álvaro J. García-Tejedor[1][¶]

[1] CEIEC Research Institute, Universidad Francisco de Vitoria, Ctra. M-515 Pozuelo-Majadahonda km. 1,800, 28223 Pozuelo de Alarcón (Madrid), Spain.

Corresponding author*
e-mail: alberto.nogales@ceiec.es

**Abstract.** The world is becoming more globalized every day and people can buy products from almost every country in the world in their local stores. Given the different food and feed safety laws from country to country, the European Union began to register in 1977 all irregularities related to traded products to ensure cross-border monitoring of information and a quick reaction when risks to public health are detected in the food chain. This information has also an enormous potential as a preventive tool, in order to warn actors involved in food safety and optimize their resources. In this paper, a set of data related to food issues was scraped and analysed with Machine Learning techniques to predict some features of future notifications, so that pre-emptive measures can be taken. The novelty of the work relies on two points: the use of categorical embeddings with Deep Learning models (Multilayer Perceptron and 1-Dimension Convolutional Neural Networks) and its application to solve the problem of predicting food issues in the European Union. The models allow several features to be predicted: product category, hazard category and finally the proper action to be taken. Results show that the system can predict these features with an accuracy ranging from 74.08% to 93.06%.

## 1    Introduction

By observing the objects that people use every day, we see they come from different places all around the world. From the chair used in the workplace, to the car used for the daily commute. This is a good demonstration of the globalized world where people live today. [1] defines globalization as the inexorable integration of markets, transportation systems, and communication systems to a degree never witnessed before. Globalization increasingly involves more people and more economic sectors every day. One of the sectors where the impact has been greatest is the transport of goods. But the safety of food and feed is a particularly sensitive issue: for consumers, food should be especially safe.

The importance of food safety can be seen in the economic impact or health risk of recent food crises. [2] shows the drop-in beef consumption after the outbreak of mad cow disease. [3] shows that disclaimers must be trustworthy, demonstrated by the disastrous mistake in detecting the origin of E. coli-contaminated cucumbers. This is a significant problem for industry: the World Trade Organization (WTO) estimates that food and farm products[1] represent approximately 10% of all exports. In fact, establishing high food safety standards is a key policy priority for the European Commission, [4]. In pursuit of this policy, the independent European Food Safety Authority (EFSA) was created in 2002 to give scientific advice and communicate the risks associated with the chain food.

EFSA, the EU Commission and European Union Member States, Iceland, Liechtenstein, and Norway are part of the Rapid Alert System for Food and Feed, RASFF[2]. The RASFF provides authorities with an effective tool to exchange information about measures taken in response to serious risks detected in food or feed. It comprises a set of records providing information about cases where health risks have been detected. By registering all notifications, a network is created that enables countries to react quickly and effectively when a health threat occurs. This network is formed by contact points working at a national level and part of the EU Commission or EFSA. These contact points are responsible for registering all information in an online system called the RASFF Portal[3]. These notifications involve many people and resources with a number of tasks: food inspectors select which products may be suspect, either in the market or at the border. Depending on the category of the selected product, different tests and analyses can be carried out, each one specific to a different hazard. If the product presents any irregularity, the corresponding authority decides how to address the problem and if the problem should be communicated to RASFF through an online form with details about the incident. This set of tasks, therefore, constitutes a simplified version of the RASFF process for the notification of food and feed issues. Each member state is operational 24/7 so urgent notifications can be made at any time. This allows officers on-duty to be forewarned and prepared to make the appropriate decision.

It is estimated that in some cases only 2% of the imported products can be analyzed[4]. An accurate prediction of products and hazards can optimize the resources available for inspections and speed up detection. For example, if the product category with more possibilities of posing a risk is predicted, resources can be focused on conducting more specific analyses.

Prediction refers to the output of an algorithm that provides a value for a future event (forecasting) or the likelihood of a particular outcome (classification). This is an intrinsically difficult task. In addition, reliable predictions require a large amount of data, this being an additional problem to the prediction itself. Furthermore, as seen in previous

---

[1] https://data.wto.org/

[2] https://ec.europa.eu/food/safety/rasff_en

[3] https://webgate.ec.europa.eu/rasff-window/portal/

[4] https://ec.europa.eu/food/sites/food/files/safety/docs/oc_leg_imports_dpe_ms_border-checks-results_2013.pdf

work [5, 6, 7] the problem of prediction in food and feed safety has not yet been studied in depth.

Machine Learning techniques are a way of identifying patterns in data after having been trained on a historical dataset and use them to automatically make predictions or decisions when applied to new data. Among Machine Learning techniques, Deep Learning has been a breakthrough in recent years, offering good results in prediction problems. Deep Learning is performed by deep neural networks, a subset of artificial neural networks inspired in how human neurons work. It is defined [8] as models which can learn representations of data with multiple levels of abstraction.

Data availability is not a problem for the work presented in this paper as the RASFF portal stores information since 1979. Data is downloaded automatically and periodically from the RASFF portal. This information is then pre-processed by using Data Mining techniques, to feed predictive models with data in the proper format. Finally, Deep Learning models have been used to predict a total of three different issue characteristics, at different stages in a simplified RASFF workflow: first, the product category that will cause the issue so authorities can focus on analyzing these products. Then, the hazard that will cause the problem in these products in order for resources to be assigned to the most appropriate analysis. Finally, the most appropriate measures to be taken to deal with that issue. By predicting the first two features, human and technical resources can be optimized which means cost savings. The last feature is directly related to avoiding wrong decision that could have an impact in the health of the population.

This paper is organized as follows. Section 2 describes the process of collecting data from RASFF, how the data has been pre-processed before being used as input to the models, and the architecture of these models. Section 3 shows the results of some experiments, the performance of the models and a comparison with the results obtained with some non-neural techniques. Finally, section 4 contains conclusions and future work.

## 2    Background

The present work uses Deep Learning techniques to make predictions in food safety by using data provided by RASFF. Previous work has shown that prediction is a typical problem of Machine Learning applied to several different fields. [9] applied bagging methods for stock market forecasting; a model based on logistics regression and applied to advertising is presented in [10]; in [11] an earthquake predictor based on AdaBoost and genetic algorithms is described; [12] uses Deep Learning architectures like LSTM and Convolutional Neural Networks (CNNs) to predict the location of cars; and finally, [13] applies DL for traffic prediction.

Prediction in the field of food safety has been done using classical Machine Learning techniques but also with DL. Fuzzy Cognitive maps are used by [14] to detect critical points in a production food chain; [15] uses a Bayesian tool for the prediction of foodborne diseases. Deep Learning is used by [16] to predict safety in meat products; [17] used a CNN trained with satellite pictures while [18] uses images to detect diseases in tomato leaves.

Information from RASFF has already been used in other works: [19] uses Network Analysis with alerts from 2003 to 2008; [20] makes a statistical analysis of issues with Listeria monocytogenes; [21] works only with data related to seafood products from 2011 to 2015; [22] also uses Network Analysis with food notifications from 2000 to 2009; [23] measures how European Union Member States contribute to the RASFF. [24] analyses only food issues related with pathogenic and non-pathogenic microorganisms; [25] uses RASFF issues with Romanian products while [26] works only with a four-year period from 2003 to 2007. Although these make use of RASFF data, they apply different approaches; none of them make predictions (in particular with DL techniques) nor use all data provided by the RASFF.

The most similar previous work to this paper is by [27], using RASFF data to predict a particular feature of health warnings using Bayesian Networks. In contrast, this work presents a Deep Learning approach, creating a training dataset using all the historical data in the RASFF to predict a set of features.

## 3 Data source and methods

This section provides some information about the materials and methods applied in the experiments. Regarding materials, we will explain how the data was obtained with an in-depth description of the data and its characteristics. All methods used are described theoretically with references given.

### 3.1 Data source

This research forecasts incoming food notifications using several Deep Learning models. These models require an adequate dataset for training and testing. Thus, the first step is data acquisition and description, followed by data pre-processing and the selection of appropriate training dataset.

**3.1.1 Downloading data from the RASFF portal.** The information is obtained from the RASFF portal, a register of food and feed issues from European Union countries, Norway, Liechtenstein, and Iceland collected since 1979. However, the RASFF portal only allows users to download the latest 5,000 items (from almost 55,000 as of 13[th] November 2019) and this research seeks to make predictions based on all stored historical data. Another problem with the dump provided by the RASFF portal is that it is an XLS file with aesthetic frames, badly defined separators and unordered records. The third problem is that downloaded information only contains basic features and a record composed of specific features tagged as details. Fig. 1 and Fig. 2 show two Web snapshots showing how information is structured. The first corresponds to the basic features of two records. The second shows the detailed information that can be accessed after querying a record.

**Fig. 1.** General information of two records on the RASFF portal.

**Fig. 2.** Detailed information in a record on the RASFF portal.

To solve these problems, Web scraping techniques were used to generate structured data based on available unstructured data on the website [28]. By using these techniques, a Web scraper was developed by coding a Python script. This script uses different libraries such as Selenium[5], used for Web testing and can be used to allow users to download information; Time[6] provides some time-related functions used to avoid a timeout causing the interruption of the script and the CSV module[7] is used to read and write CSV files. This will create a CSV file containing all the historical records that can be updated at any time. In the CSV, each row will correspond to a registered issue and each column corresponds to a general or detailed feature. All features are scraped except for some specific features: last update published in the RASFF, analytical results, unit and sampling date.

The flowchart of the scraper is represented in Fig. 3. The RASFF portal URL shows pages with 100 records along with the general information mentioned above. The first function of the script accesses this URL iteratively with a second function iterates these 100 records, saving the univocal identifier that will be then used to access to the detailed information of the record and all the general features contained in that record. A set of functionalities is responsible for saving the detailed features of each record and to verify all information is correctly extracted. Finally, there is also a set of functions that pre-process the data: deleting blank spaces that are not useful or characters that can produce errors due to its format.

**Fig. 3.** Web scraper flowchart.

**3.1.2 Data description**. At this point, the result after scraping is a CSV file containing all the records in the RASFF portal since 1979, with almost all the features for each one. It will be ordered by date (starting with the most recent) and following a tabular structure so it can be easily accessed by the basic Python modules. All features except two correspond to categorical variables written in English. The following is a brief description of each feature and its possible values:

- NUMBER contains an identifier assigned to each record identifying its order in relation with how recent it is. The most recent record will be identified by

---

[5] https://selenium-python.readthedocs.io/index.html

[6] https://docs.python.org/2/library/time.html

[7] https://docs.python.org/3/library/csv.html

1. This number changes when a new issue is registered. This feature will not be included in the final dataset.

- CLASSIFICATION contains a classification entered by RASFF administrators based on the reporting of safety issues. It has 5 different categories: alert, information for follow-up, border rejection, information for attention and information.
- DATE_CASE corresponds to the day the issue takes place. It follows the format dd/mm/yyyy. It has a cardinality of 12.
- REF contains the code used to identify the record in the RASFF system. With this code the general and detailed features of each record can be searched directly in the RASFF portal without using the forms of the Website. This feature will not be included in the final dataset.
- NOTIFICATION_COUNTRY, this feature gives the name of the country registering the issue. As indicated above, only European Union countries, Norway, Liechtenstein, and Iceland can perform this action. The cardinality is 32.
- SUBJECT, it contains a free summary limited to 200 characters introduced by administrators from the country with the safety issue.
- PRODUCT_CATEGORY, this feature is used to identify the category of the product analyzed. It has 38 possible options, such as nuts, fruits, and vegetables or milk and milk products.
- TYPE, this is a categorization of products within a hierarchy. There are three options: food, feed and food contact material (fcm).
- RISK_DECISION, this is used to evaluate the notification about the importance of the issue. It could be serious, not serious and undecided.
- ACTION_TAKEN, this refers to the action carried out by each country. It can have 24 possibilities like informing authorities, re-dispatch or destruction.
- DISTRIBUTION_STATUS, the information given by this feature is related to how the problem product is distributed at that moment. It could have 17 different values, such as, no distribution to other member countries or product already consumed.
- PRODUCT, this feature gives the specific name of the product. For example, groundnuts.
- HAZARD, this identifies the hazards or anomalies that have caused the issue (it could be one or various). For example, undeclared peanut or sulphite content.
- HAZARD_CATEGORY, this contains the categories where each of the hazards or anomalies from the previous variable are classified. Values may refer to allergens or food additives and flavourings with a cardinality of 35.
- COUNTRY_ORIGIN, this identifies the country of origin of the product. This could be any country in the world. Its cardinality goes to 190.
- COUNTRY_DESTINATION, this is the destination country or countries of the product. This also could be any country in the world. In this case, cardinality is 215.

- COUNTRY_DISTRIBUTION, this indicates the countries within the transportation chain of the product, except the origin and destination country. That is, the countries through which the product is en route to its final destination. This could be any country in the world or certain international regulatory organisations. It has a cardinality of 138.

As it can be seen, the scraped dataset is a compound of mostly categorical variables; this should be considered as it will be key to the way data is treated before applying Deep Learning techniques.

**3.1.3 General pre-processing**. Once the dataset has been downloaded, a general pre-processing for cleaning and formatting is carried out before realizing the specific coding for each model. This pre-processing uses classical Python libraries like pandas or NumPy, pandas[8] and NumPy[9]. This process cannot be done during the scraping stage as a general vision of all the information is required. The following tasks are accomplished in this stage:

- String replacement in empty texts and NaN values for a string with a blank value. This is done because these values can cause an error when using the encoding techniques that each model needs.
- Category formatting. There are cases in which the same categorical value has been written in different ways. For example, when the value has been written with or without capital letters.
- Duplicity removal. As a great deal of time may be needed to scrape all data from the website, new issues may be registered. This could cause some duplicates that must be deleted.
- Category renaming. As the RASFF has been in operation for four decades, the names of some categories have changed. In order to avoid these being treated as different, outdated categories are updated to the current values.
- Format change in some variables. For faster and more flexible data processing, some datatypes have been formatted. For example, DATE_CASE has been converted from string to DateTime.

**3.1.4 Choice of the final dataset.** In order to analyze distribution over the years, scraped data has been plotted chronologically with Matplotlib[10], a graphics library. As can be seen in Fig 4, 50,416 records out of 56,385 were produced from 2004 to 2019 at a rate of some 3,000 food records per year. This is 88.93% of the total, that is, before 2004 the number of issues was low (about 100 records per year in the first period and 500 in the final). This difference of incidents per year can lead to biases in predictions and these residual records can cause more inconveniences than benefits when training the models. For this reason, the final dataset will only contain 47,981 records from 2004 to 2019.

---

**Fig. 4.** Records per year registered in RASFF.

A second transformation on the dataset is related to the way Machine Learning/Deep Learning models are trained. Training can lead to overfitting if it maximizes the model's performance on the training set and performs poorly on unseen data. The model then begins to memorize training data rather than learning to generalize from them. Thus, a common practice is to split the dataset and keep part of it as a "validation set" on which the training performance is measured. However, the model can still be too adapted to the validation set if the hyperparameters are modified excessively, looking for optimal performance. In this case, evaluation metrics will not represent the quality of the generalization. To solve this problem, the original data set is split again, and a new test set is generated. The training continues in the training set, after which an evaluation is made with the validation set. When the experiment seems to be successful, the final evaluation is done in the test set.

The experiments conducted for this work used data from 2004 to 2018 as the training and validation set. These were randomly split into 80% and 20% respectively. We have preserved 2019 data for testing. This decision was based on the time series and annual seasonality of food and feed alerts. The testing instances were not seen by the models during the training stage and are entirely new to them. Table 1 shows the number of records in each dataset division.

**Table 1.** Number of records used at each stage.

| Dataset | Records |
|---|---|
| Train | 38,102 |
| Validation | 9,525 |
| Test | 2,789 |
| Total | 50,416 |

By dividing the available data into three sets, the number of samples that can be used is drastically reduced and the results may depend on a particular random choice for the training/validation sets. To avoid this problem, k-fold cross-validation was used. The training set is divided into k smaller sets or "folds" and a loop is carried out for each of the k folds: the model is trained using k-1 folds as training data and validated (measuring its performance) with the remaining fold. The performance measure reported by the cross-validation is the average of the values calculated for the whole loop.

### 3.2    Methods

Once the dataset has been obtained and transformed, few methods need to be applied; the first group is mostly related to the nature of the data as in this case all features are categorical. The second group consists of the neural models used for predictions and a set of non-neural models used to validate the proposed ones.

**3.2.1 Encoding of categorical variables.** Most of the Machine Learning algorithms cannot work with categorical data. They require all the inputs and outputs to be numerical and therefore all variable in the dataset must be encoded. In other words, to convert this data into numbers while preserving as far as possible the information within the dataset. The choice of encoding is directly related to the accuracy of the model. In this case, this impact can be even greater as all variables in the dataset are categorical. Therefore, the models proposed in this paper are the sum of the encoding and the Machine Learning model. The different types of proposed encodings are subsequently introduced.

*Integer encoding.* This is the simplest form of encoding. In this case, every category of a categorical variable is transformed into integers. Number 1 being given to the first category, 2 for the second and so on, till *n* which is the number of different values that the categorical variables can take. An example of this encoding could be TYPE which can take 3 values, as shown in Table 2.

**Table 2.** Integer encoding for variable TYPE.

| *Original data* | | *Encoded data* |
|---|---|---|
| **ID** | **TYPE** | **TYPE** |
| 0 | food | 1 |
| 1 | feed | 2 |
| 2 | fcm | 3 |

*Binary encoding.* This consists of converting into a binary the values obtained with another encoding method. By doing so, a new variable is created for each of the binary digits necessary to represent the total number of possible categories. The number of variables needed for encoding is calculated by applying Equation 1. It considers *n*, the number of possible values of the categorical variables. The equation uses the ceil function that calculates the smallest integer greater or equal to a given number. An example of this encoding is shown in Table 3

$$ceil(log_2(n+1)) \qquad (1)$$

**Table 3.** Binary encoding for variable TYPE.

| *Original data* | | *Encoded data* | |
|---|---|---|---|
| **ID** | **TYPE** | **TYPE-0** | **TYPE-1** |
| 0 | Food | 0 | 1 |
| 1 | Feed | 1 | 0 |
| 2 | Fcm | 1 | 1 |

*Feature hashing.* This encoding makes use of the hashing trick, consisting of creating a hash table using a function of the same name. It transforms input elements or strings of any length into output, a numerical code of a fixed length determined by the function, as depicted in Equations 2 and 3.

$$H: U \to M \qquad (2)$$

$$x \to h(x) \qquad (3)$$

Function hash ($H$) is the projection of set $U$ (set of data to be codified) over the set $M$ (set of codified data). An element of $M$ ($h(x)$) is calculated by applying the hash function to an element $x$ from $U$.

*One hot encoding.* This is one of the most common methods of encoding categorical variables. In this case, each value of the categorical variable is represented by a vector of size $n$. The size is given by the possible values of the variable. The vector will have all the positions with value 0, except one with value 1 representing the value that is being coded. This encoding is obtained using Equation 4, $\nu$ being the vectors that represent the encoding for each of the categories of a categorical variable of $n$ categories. An example is provided in Table 4.

$$\nu \in \{0,1\}^n : \sum_{i=1}^{n} v_i = 1 \qquad (4)$$

**Table 4.** One Hot encoding for variable TYPE.

| Original data | | Encoded data | | |
|---|---|---|---|---|
| **ID** | **TYPE** | **TYPE-food** | **TYPE-feed** | **TYPE-fcm** |
| 0 | food | 1 | 0 | 0 |
| 1 | feed | 0 | 1 | 0 |
| 2 | fcm | 0 | 0 | 1 |

By using this encoding, the encoded variable that consists of $x$ instances and $n$ possible values will be transformed into $n$ binary variables each with $x$ instances. In each instance, it will be indicated the presence or absence with 1 or 0. Therefore, the number of variables in the dataset will grow similar to the binary and hashing encoding. The main advantage of this type of encoding is that it provides equidistant representation of all categories and does not assume arbitrary relationships between them, such as integer encoding. This however is one of its disadvantages. Categories often resemble each other or share some characteristics. To solve this problem and provide a more faithful representation, the technique indicated below can be used

*Entity Embedding of Categorical Variables.* [29] proposes this method that reduces the use of memory and accelerates the process of model formation compared to One Hot Encoding. This type of encoding performs well with Neural Network models.

For multi-dimensional spaces of categorical features, this method automatically maps closer categories with similar effect to the target output, thus helping neural networks to solve the problem. In other words, entity embeddings are used to map categories into a continuous distributed vector in a semantic space. In this space, similar categories are closer. More interestingly, the distance between categories is meaningful but also the direction of the vectors. This allows intrinsic properties of categorical variables to be identified.

This coding is performed using embedding layers. Each categorical variable is an input to the model and needs a different inlay. That is, the relationships sought are those existing between the category values of each categorical variable, not the relationships that exist between variables. The output dimension of each of these layers is a hyperparameter that will be modified during the training process considering the expected output value of the training data. Fig. 5 shows how embedding layers are connected to the neural model.

**Fig. 5.** Example of neural model with embedding layers.

The operation of an embedding layer can be understood as a table in which each different entry (each category) has an associated index that relates it to an embedding vector of a pre-established dimension. Therefore, each category has a unique vector associated with it. The values of these vectors are randomly initialized and modified as if they were in a neural network, depending on the final error made by the prediction. Thus, the representation of the categories made by a model that tries to solve a specific problem will foreseeably be a representation that conserves only the relevant characteristics and relationships when solving that specific problem.

The output vectors of each of the embedding layers are concatenated to each other forming a vector that represents the complete codification of the input, being the union of the codifications of all the variables that compose this input. This vector will, therefore, be the input of the model that follows the coding.

**3.2.2 Neural models.** The experiments described in this paper apply deep neural networks models to RASFF portal data to predict three different issue characteristics, at different stages of a simplified RASFF workflow. The approach is a sequential chain where inputs for a stage are the same as the previous stage plus its output. Thus, three different predictors were developed, each one built with the required encodings and a particular neural architecture. Although several neural models were tested, the best results were obtained with Multilayer Perceptrons (MLP) and 1D Convolutional Networks (CNN).

*Multilayer Perceptron (MLP)* is a feedforward supervised neural network model. It consists of an input layer, an output layer and an arbitrary number of hidden layers. The basic MLP has a single hidden layer. Neurons use non-lineal activation functions, either sigmoid, hyperbolic tangent or Rectified Linear Unit (ReLU). Learning is carried out through backpropagation using the Generalized Delta Rule to update the weight matrices.

*1D Convolutional Neural Network (CNN)* is a type of feedforward supervised deep neural network that is able to model high-level data abstractions using hierarchical architectures. They learn input-output relationships based on convolution operations over a one-dimensional array. Each convolutional layer extracts hierarchically and incrementally some characteristic of the input array. They have also been used with texts, so it makes sense to test its performance in instances with categorical features.

**3.2.3 Optimization of model hyperparameters.** The behaviour of Deep Neural Networks is controlled by different hyperparameters, based on a trial and error basis. This is a time-consuming task, so a method called grid search is used to find an optimal combination. Grid search is a method that combines a set of hyperparameters with different values, obtaining the model with the best accuracy [30].

**3.2.4 Food and feed prediction system.** The main aim of this research is to predict a series of features related to the simplified RASFF system previously described. This will provide the different roles involved with extra information optimising decision making. After a thorough investigation, it was concluded that RASFF can take advantage of this work in three key points. These are defined as the three stages of the workflow proposed below:

- The moment border officials or inspectors decide which products could be contaminated. A series of products having different characteristics arrive from different countries. Due to the limited amount of human resources, not all the products can be analyzed. By predicting which products have a higher probability of being contaminated, they can focus only on taking samples of a few products. Thus, at this stage, the feature PRODUCT_CATEGORY is predicted.
- In the laboratory, once the product to be analyzed has been identified. A single product can present multiple hazards, each one requiring a different type of analysis. At this point, there is an interest in predicting the hazard that can be found in the product, so the usage of the laboratory equipment is optimized. In this case, the feature to be predicted is HAZARD_CATEGORY.
- Finally, once a contaminated product is found, a decision must be made about what to do. The variable predicted at this stage is ACTION_TAKEN.

Fig 6 depicts the stages of the simplified RASFF system and in parallel the proposed workflow of the developed system with the categories that are predicted.

**Fig. 6.** Simplified RASFF and stages of the prediction system.

As the cases are sequential, the inputs of a predictor will be the inputs of the previous one plus the prediction made at the previous stage. Table 5 shows which features are the inputs for each stage, and the range of different values that the predicting variable can take.

**Table 5.** Inputs and outputs of the models at different stages.

| Stage | Inputs (different categories with different values) | Output (one category with different values) |
|---|---|---|
| 1 | Date, Notification Country, Distribution State, and Origin Country | Product Category: 38 different values |
| 2 | Inputs of stage 1 plus Product Category | Hazard Category: 35 different values |
| 3 | Inputs of stage 2 plus Hazard Category | Decision Taken: 24 different values |

**3.2.5 Non-neural models.** As part of the validation of the proposed models, some baseline models were used in order to evaluate their accuracy. They were also used with the different proposed encoding except for categorical embeddings as they can only be used with neural models. These models were developed with the help of Python library scikit-learn[11] used for Data Mining.

*Logistic regression.* This is a technique that belongs to what are called linear generalized models. The main characteristic of this model is that it is able to predict a qualitative variable based on several predictive ones. According to [31], it analyzes the relationship between multiple independent variables and a categorical dependent variable.

*Decision trees.* This model consists of predicting the target variable by learning simple decision rules inferred from the data features. It is defined in [32] as: "a classifier expressed as a recursive partition of the input space based on the values of the attributes". It builds logic diagrams with the form of hierarchical trees. They represent the categorization of the data under a series of conditions applied in the form of successive. In the tree, each node represents a test or decision on an attribute, and each branch the result of the test and each node completes a class label.

*Random forest.* Introduced by [33], this is a method based on decision trees. The technique consists of combining them and averaging the models to improve the results. Compared with decision trees, this technique usually reduces the problem of overfitting.

*Boosting trees.* This technique is also based on decision trees and is defined in [34] as: "gradient boosting trees are tree ensemble methods that build one decision tree learner at a time by fitting the gradients of the residuals of the previously constructed tree learners". It makes use of the different techniques that optimize and improves gradient descent over the loss function.

*Support Vector Machine (SVM).* The actual version of SVM was proposed by [35]. It can be seen as a classifier where instances are distributed through an n-dimensional space. The objective of the algorithm is to find a hyperplane that divides individuals into different classes making the separation between them as wide as possible.

# 4    Results

## 4.1    Selection of the neural model

Some hyperparameters can be decided at the beginning as they will remain fixed in the different model configurations. Since the project deals with categorical variables, output layers for both MLP and CNN models are built with as many neurons as categories that must be predicted at each stage. The activation function for these output layers is Softmax, which produces a value between 0 and 1 for each output neuron. That value is the probability that this neuron represents the correct output of the network. The sum of the outputs of all the neurons in the output layer is equal to 1. The loss function selected was categorical cross-entropy since the problem can be defined as label-categorization task.

---

[11] https://scikit-learn.org/stable/

All models were developed using Keras[12], a high-level API built on top of TensorFlow[13], the Google open-source library for Machine Learning and Deep Neural networks.

Different neural models and encodings were tested to select the most suitable combination for each prediction. Two models (a Multilayer Perceptron and a 1D Convolutional Neural Network) were developed and trained with a combination of classical encodings and Entity Embedding of Categorical Variables. Neural architectures were designed and tuned using iterative grid search techniques applied to the two possible baseline models (MLP and Conv1D) with One Hot encoding (standard coding). A first iteration was used to select the number of neurons, hidden layers and the activation function for each layer. From these results, the dropout between different hidden layers was fixed. In a new iteration, the number of epochs and optimizers were selected. The last hyperparameter that was set was the optimal learning rate. For the Conv1D model, the same process was followed except in the first iteration, where the number of neurons is changed for the number of filters, kernel sizes and Maxpooling. Tables 6 and 7 show the hyperparameters that were tested and their different values for the MLP and the Conv1D, respectively. The final configuration of the three models is shown in Appendix: Fig. 1, 2 and 3.

**Table 6.** Configuration of grid search for MLP (Hyperparameters and values).

| Stage | Hyperparameter | Values |
|---|---|---|
| | Hidden layers | 1, 2, 3, 4 |
| 1 | Neurons per layer | 128, 256, 512, 1024, 2048, 4096 |
| | Activation function | ReLU, tanh, Sigmoid, Hard Sigmoid |
| 2 | Dropout | 0.1, 0.2, 0.3, 0.5 |
| | Epoch | 10, 25, 50, 100, 150 |
| 3 | Optimizer | Stochastic Gradient Descent, Root Mean Square Prop, AdaGrad, Adam |
| 4 | Learning Rate | 0.01, 0.001, 0.005 |

**Table 7.** Configuration of grid search for Conv1D (Hyperparameters and values).

| Stage | Hyperparameter | Values |
|---|---|---|
| | Hidden layers | 1, 2, 3, 4 |
| | Number of filters | 32, 64, 128, 256, 512 |
| 1 | Kernel size | 2, 3, 4, 5 |
| | MaxPooling | 2, 3, 4 |
| | Activation function | ReLU, tanh, Sigmoid, Hard Sigmoid |
| 2 | Dropout | 0.1, 0.2, 0.3, 0.5 |
| | Epoch | 10, 25, 50, 100, 150 |
| 3 | Optimizer | Stochastic Gradient Descent, Root Mean Square Prop, AdaGrad, Adam |

---

[12] https://keras.io/

[13] https://www.tensorflow.org/

| 4 | Learning Rate | 0.01, 0.001, 0.005 |
|---|---|---|

Tables 8, 9 and 10 with both Deep Learning Models when input data was coded with One Hot Encoding (the best results among classical encodings) and Entity Embeddings of Categorical Variables. The experiments were carried out five times each, starting each time with randomly generated weight matrices. The results are the average of the five runs, thus avoiding deviations from the mean that could be obtained with a single experiment. The accuracy is measured in the output layer of each model as the number of times the most likely category matches the correct one.

**Table 8.** Accuracy of the different models. Stage 1.

| Model | One Hot Encoding | Entity Embedding of Categorical Values |
|---|---|---|
| MLP | 69.67 % | **72.25 %** |
| 1D CNN | 69.13 % | 70.12 % |

**Table 9.** Accuracy of the different models. Stage 2.

| Model | One Hot Encoding | Entity Embedding of Categorical Values |
|---|---|---|
| MLP | 78.57 % | 80.92 % |
| 1D CNN | 78.18 % | **81.83 %** |

**Table 10.** Accuracy of the different models. Stage 3.

| Model | One Hot Encoding | Entity Embedding of Categorical Values |
|---|---|---|
| MLP | 80.35 % | **81.26 %** |
| 1D CNN | 77.74 % | 78.53 % |

Best accuracies in stage 1 and 3 are obtained with an MLP with Entity Embedding of Categorical Values. For stage 2, the best result is achieved with the same encoding but using a 1D Convolutional Neural Network.

Stage 1 model is formed by four embedding layers of size 6, 16, 9 and 50 that are concatenated with an MLP with three hidden layers of 2048, 1024 and 512 neurons. The activation function for these layers is ReLU (Rectified Linear Unit), which represents a nearly linear function and therefore preserves the properties of linear models that made them easy to optimize, with the gradient descent method, [37]. The output layer has 38 neurons that correspond to the product categories. Error is measured with the categorical cross-entropy and Adam has been used as the optimizer.

Stage 2 model uses one more embedding layer of 19 neurons. These layers are concatenated with a 1D Convolutional Neural Network of two-layer of 512 and 256 neurons. The number of the convolutional filters is 128 and 256 with size 4 and 3 respectively. The output layer has 35 neurons (number of hazard categories).

Finally, stage 3 model is similar to model 1 with two differences: it has two extra embedding layers with 19 and 18 neurons and the output has 24 neurons that correspond to the different actions to be taken.

## 4.2    Probabilistic predictions

As mentioned above, the output in the three models is a vector with the size of the number of neurons in the output layer. The value at each position goes from 0.0 to 1.0, summing up in total 1.0. The meaning of each position is the probability of this value to occur.

Non-neuronal models make a single bet on what the expected category will be. Neural models with a probabilistic result can be used to reduce the scope of products, hazards or actions to a small number of options (not just one). This broadens the scope of the search but still guarantees a higher percentage of success. Following this approach, three different predictions have been made: Top1, Top2, and Top3.

- Top1. This checks if the category to which the network gives the greatest probability matches the real one. It is the accuracy used to measure the models in Tables 8, 9 and 10.
- Top2. In this case, the accuracy is calculated by checking if the actual category is among the two categories to which the network has given the most probabilities and adding both accuracies.
- Top3. In this case, the accuracy is calculated by checking if the actual category is among the three categories to which the network has given the most probabilities and adding the accuracies in the three categories.

Table 11 summarizes the results obtained at each stage with the same models but calculating accuracies as described above. Top3 models logically improve accuracy.

**Table 11.** Accuracies of the three models depending on the type of prediction.

| Stage | Top1 | Top2 | Top3 |
|---|---|---|---|
| 1 | 72.25 % | 82.94 % | 84.88 % |
| 2 | 81.83 % | 91.27 % | 93.53 % |
| 3 | 81.26 % | 88.99 % | 89.76 % |

## 4.3    Evaluation against non-neural models

At this point in the study, the proposed architectures with neural models have performed quite well. Therefore, they should be validated against baseline models of a non-neural nature. Previously, we defined a set of different models: logistic regression, decision trees, random forest, boosted trees and SVM. These architectures along with the different proposed encodings are combined,  creating new models to be compared with the chosen neural architectures from the previous subsection. Results of accuracy can be seen in Table 12, 13 and 14, again one table for each stage. In the case of neural models, the accuracy corresponds to Top1 prediction.

**Table 12.** Evaluation against non-neural models in stage 1.

| Model | Integer | Binary | Hashing | OHE | Categorical Embeddings |
|---|---|---|---|---|---|
| Logistic regression | 17.00 % | 43.52 % | 44.48 % | 51.12 % | - |
| Decision trees | 68.85 % | 68.70 % | 66.40 % | 69.13 % | - |
| Random forest | 68.19 % | 67.07 % | 67.45 % | **69.55** % | - |
| Boosted tree | 68.12 % | 69.31 % | 68.05 % | 69.38 % | - |
| SVM | 33.54 % | 40.02 % | 43.78 % | 43.15 % | - |
| MLP | 43.86 % | 64.06 % | 69.53 % | 70.44 % | **72.25** % |

From all the combinations of non-neural architectures, decision trees with One Hot Encoding is the one that performs better. It obtains an accuracy of around 70%. MLP with categorical embedding performs 3% better.

**Table 13.** Evaluation against non-neural models in stage 2.

| Model | Integer | Binary | Hashing | OHE | Categorical Embeddings |
|---|---|---|---|---|---|
| Logistic regression | 22.10 % | 54.89 % | 54.77 % | 58.21 % | - |
| Decision trees | 78.12 % | 77.65 % | 77.88 % | 80.33 % | - |
| Random forest | 79.55 % | 79.12 % | 75.76 % | **80.89** % | - |
| Boosted tree | 78.99 % | 80.15 % | 78.30 % | 80.80 % | - |
| SVM | 34.34 % | 51.02 % | 52.09 % | 59.02 % | - |
| Conv 1D | 56.66 % | 76.92 % | 77.93 % | 79.69 % | **81.83** % |

Regarding Table 13, the best results are with random forest with One Hot Encoding with around 80% of accuracy. The neural architecture performs almost 1% better.

**Table 14.** Evaluation against non-neural models in stage 3.

| Model | Integer | Binary | Hashing | OHE | Categorical Embeddings |
|---|---|---|---|---|---|
| Logistic regression | 28.42 % | 50.19 % | 49.09 % | 53.67 % | - |
| Decision trees | 79.59 % | 78.87 % | 76.98 % | 79.97 % | - |
| Random forest | 79.00 % | 77.81 % | 77.03 % | **80.02** % | - |
| Boosted tree | 79.77 % | 79.72 % | 77.54 % | 80.21 % | - |
| SVM | 42.75 % | 46.76 % | 48.12 % | 58.88 % | - |
| MLP | 58.48 % | 74.79 % | 79.33 % | 79.69 % | **81.26** % |

At stage 3, the best results are provided again by random forest and One Hot Encoding (around 80% of accuracy). In this case, the neural architecture performs only slightly better with a 1% improvement.

### 4.4     Evaluation of the prediction

Finally, all selected neural models were evaluated against the test dataset that was previously separated form training/validation data. For that purpose, all 2,789 instances from 2019 dataset that were never fed into the model were used. The prediction was carried out for the three stages and accuracy, calculated by the different types: Top1, Top2, and Top3. All this information is summarized in Table 15. As it can be seen accuracies are very similar to values in training stage (Table 12) which means there is no overfitting in the models.

**Table 15.** Different test accuracies for each model.

| Stage | Top1 | Top2 | Top3 |
|---|---|---|---|
| 1 | 72.05 % | 82.32 % | 84.84 % |
| 2 | 80.38 % | 89.78 % | 92.40 % |
| 3 | 81.49 % | 89.98 % | 90.90 % |

## 5     Conclusions and future works

The experiments described in this paper provide a method to forecast from food and feed safety issues using data obtained from the RASFF portal and Deep Learning architectures. Three predictive models are proposed, each of them yielding an intermediate result in the workflow of a simplified RASFF system. When all models are sequentially applied in a pipeline, the product and hazard as well as preventive action can be foreseen.

First, a scraper was developed to obtain all the information stored in the RASFF portal. Then simple statistical operations were applied to select useful data and identify relationships between different variables of the dataset. Finally, neural models were designed and implemented. This step was divided into two phases: encoding of input variables and selection of neural architecture. For the first, a specific encoding method based on entity embedding for categorical variables was used as input data are categorical entities. For the second, the most adequate neural model for each workflow stage was selected from several possible architectures (Multilayer Perceptrons and Convolutional Neural Networks).

As the output of the neural architecture is a vector with as many values as categories to be predicted at each stage, the accuracy of each model was validated in three different scenarios: a prediction based on the most likely category (Top1 prediction) and two predictions where accuracy is measured in the two (or three) most likely categories. The results thus obtained were validated in two ways: by comparing the Top1 accuracies against a set of non-neural models and by predicting 100 RASFF instances never seen by the models.

Some conclusions can be extracted from this work. It has been demonstrated that the best results are obtained by encoding data with Categorical Embeddings and using Deep Neural models for forecasting. When considering the Top1 prediction, the models perform slightly better than any approach based on non-neural architectures. This performance is improved when the second and the third most probable prediction (Top2 and Top 3 predictions) are also considered. It should also be noted that the models were tested with data never used in the training/validation phase. As can be seen, the models are robust, since the accuracies obtained are more or less the same, with no significant deviations compared to the results obtained in the validation stage.

For future work, some ideas can be highlighted:

- Forecasting food and feed problems through a time series approach. For example, count the weekly or monthly number of issues that occur for each type and, therefore, obtain several numerical time series. From this it is possible to predict when and how many instances of each type will occur at a given time.
- Take advantage of the variable summary of each record. This variable consists of a text written by the health authorities. The objective would be to obtain information using text processing techniques.
- Study the flow of food through different countries, using techniques based on graph theory.

**S1 Appendix Deep Learning architectures**